\documentclass{article}
\usepackage{ijcai26}

\usepackage{times}
\usepackage{soul}
\usepackage{url}
\usepackage[hidelinks]{hyperref}
\usepackage[utf8]{inputenc}
\usepackage[small]{caption}
\usepackage{graphicx}
\usepackage{amsmath}
\usepackage{amsthm}
\usepackage{booktabs}
\usepackage{times}
\usepackage{algorithm}
\usepackage{algorithmic}
\usepackage{multirow}
\usepackage{xcolor}
\usepackage{listings}

\usepackage[switch]{lineno}

\title{Deferred Commitment Decoding for Diffusion Language Models}

\author{
Yingte Shu$^1$
\and
Yuchuan Tian$^1$\and
Chao Xu$^1$\and
Yunhe Wang$^2$\and
Hanting Chen$^2$\\
\affiliations
$^1$Peking University\\
$^2$Huawei Technologies\\
\emails
ytshu25@stu.pku.edu.cn,
yuchuan.tian@outlook.com,
xuchao@cis.pku.edu.cn,
yunhe.wang@huawei.com,
chenhanting@huawei.com
}

\begin{document}

\maketitle

\begin{abstract}
  Diffusion language models (DLMs) have recently emerged as a strong alternative to autoregressive models by enabling parallel text generation. To improve inference efficiency and KV-cache compatibility, prior work commonly adopts block-based diffusion, decoding tokens block by block. However, this paradigm suffers from a structural limitation that we term \textbf{Boundary-Induced Context Truncation (BICT)}: undecoded tokens near block boundaries are forced to commit without access to nearby future context, even when such context could substantially reduce uncertainty. This limitation degrades decoding certainty and generation quality, especially for tasks requiring precise reasoning, such as mathematical problem solving and code generation. We propose \textbf{Deferred Commitment Decoding (DCD)}, a novel, training-free decoding strategy that mitigates this issue. DCD maintains a certainty-aware sliding window over masked tokens, resolving low-uncertainty tokens early while deferring high-uncertainty tokens until sufficient contextual evidence becomes available. Extensive experiments across multiple diffusion language models, benchmarks, and caching configurations show that DCD improves generation accuracy by 1.73\% with comparable time on average compared to fixed block-based diffusion methods, with the most significant improvement reaching 16.5\%. These results demonstrate that deferring token commitment based on uncertainty is a simple yet effective principle for improving both the quality and efficiency of diffusion language model decoding. Code: \url{https://github.com/shuyingte/DCD}
\end{abstract}

\begin{figure}[t]
    \includegraphics[width=\columnwidth]{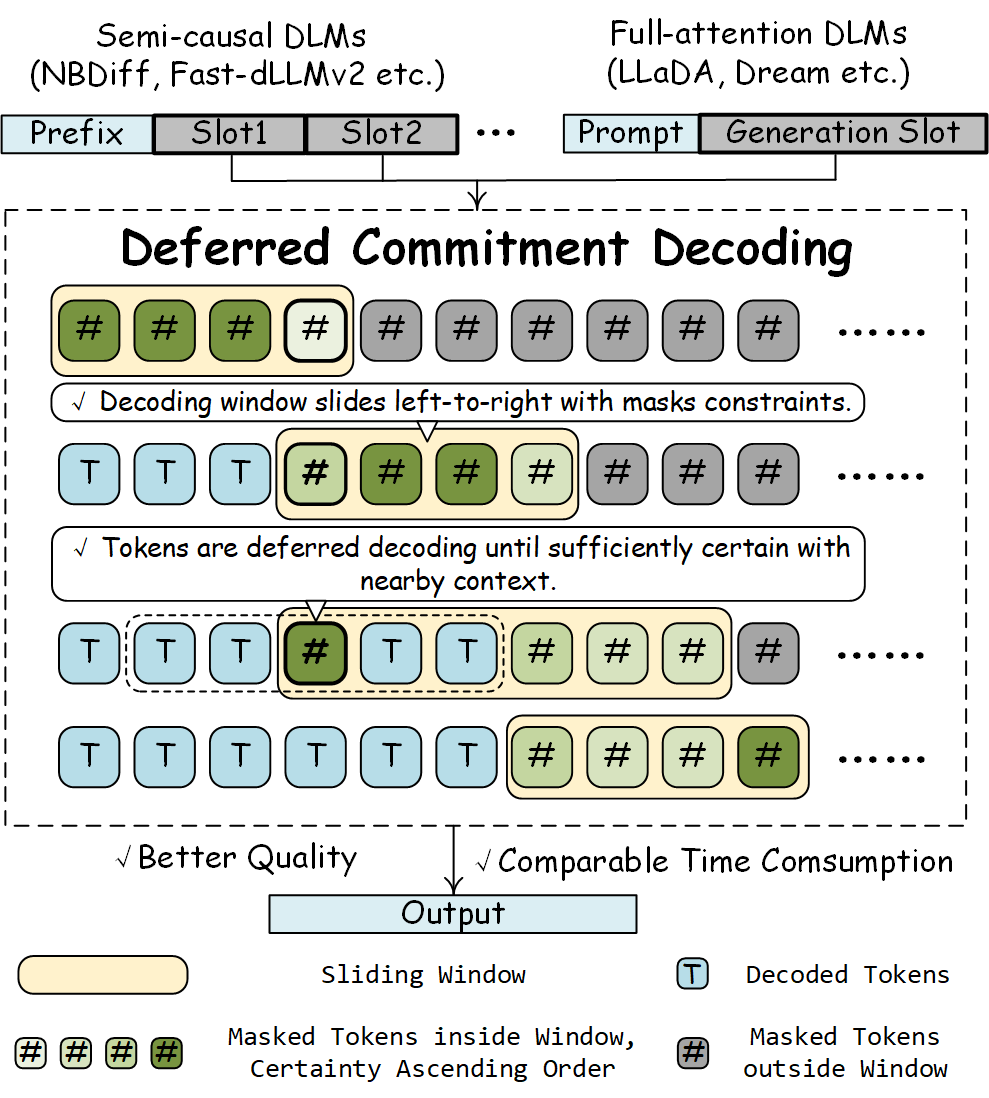}
    \caption{\textbf{An overview of the proposed DCD algorithm.} DCD defers uncertain tokens' decoding with sliding windows for DLMs, thereby improving performance.}
    \label{fig:overview}
\end{figure}

\section{Introduction}
Diffusion language models (DLMs) have recently emerged as a promising alternative to autoregressive models for natural language generation. By decoding tokens in parallel rather than strictly left-to-right, DLMs relax sequential dependencies and enable more flexible generation. Recent models such as NBDiff \cite{tian2025next} and LLaDA2.0 \cite{bie2025llada2} demonstrate that, at comparable scales, DLMs can match their autoregressive counterparts on selected reasoning tasks.

A major challenge in practical DLM inference lies in compatibility with key-value (KV) caching. Vanilla DLMs decode tokens in largely unconstrained orders, which prevents effective reuse of cached attention states and leads to slow inference. To address this issue, block-based diffusion methods have been proposed \cite{arriola2025block,nie2025large}, partitioning the sequence into blocks that are decoded sequentially while allowing parallel decoding within each block. This semi-autoregressive structure significantly improves KV-cache efficiency and has become a standard design choice in recent DLM systems.

Despite their efficiency benefits, block-based diffusion methods introduce a fundamental limitation, which we refer to as \textbf{Boundary-Induced Context Truncation (BICT)}. Once decoding proceeds to the next block, undecoded tokens in the current block are forced to commit, even if nearby future tokens—often only a few positions away—could provide crucial disambiguating context. This issue is particularly detrimental for tokens in semantically critical positions, where insufficient context leads to low-certainty decisions and error propagation. Importantly, this limitation is not caused by an incorrect decoding order but by rigid block boundaries that assume information sufficiency upon block completion.

Our core hypothesis is that decoding quality can be improved by deferring commitment on high-uncertainty tokens until sufficient contextual evidence becomes available, without abandoning the efficiency advantages of block-based decoding. Based on this insight, we propose \textbf{Deferred Commitment Decoding (DCD)}, a training-free decoding strategy that replaces fixed block boundaries with a certainty-aware sliding window. Within this window, tokens with low uncertainty are resolved first, while high-uncertainty tokens remain masked and continue to benefit from dynamically expanding context. This mechanism enables localized bidirectional information flow while preserving compatibility with existing caching schemes.

We evaluate DCD on a diverse set of tasks, including mathematical reasoning \cite{math500,gsm8k}, code generation \cite{mbpp,humaneval}, and instruction following \cite{ifeval}, using multiple diffusion language models \cite{nie2025large,ye2025dream,wu2025fast2,tian2025next} and various KV caching configurations. Across all settings, DCD consistently improves generation accuracy by \textbf{+1.73\%} with comparable inference time on average compared to fixed block-based diffusion baselines, with the maximum improvement in certain configurations reaching \textbf{+16.5\%}. These results establish DCD as a strong state-of-the-art decoding method for DLMs.

We summarize our contributions as follows:
\begin{itemize}
\item We identify \textbf{Boundary-Induced Context Truncation} as a key structural limitation of block-based diffusion decoding, which prevents undecoded tokens from leveraging nearby future context across rigid block boundaries.
\item We propose \textbf{Deferred Commitment Decoding}, a simple, training-free decoding strategy that dynamically aligns the decoding order with token-level uncertainty using a sliding window.
\item We demonstrate that DCD achieves consistent accuracy improvements over fixed block-based diffusion methods across models, tasks, and caching configurations.
\end{itemize}

\section{Related Works}
\subsection{DLMs Taxonomy}
There are two main lines of work that adapt diffusion techniques \cite{ho2020denoising} from computer vision to natural language processing. \emph{Continuous diffusion language models} \cite{li2022diffusion,gong2022diffuseq} project discrete language tokens into continuous spaces and apply denoising processes to recover text outputs. In contrast, \emph{discrete diffusion language models} draw inspiration from masked language modeling \cite{devlin2019bert}, gradually recovering masked tokens at predefined generation slots. Compared to continuous approaches, discrete DLMs better align with the inherently discrete nature of language and can be more easily adapted from existing autoregressive models \cite{gong2024scaling}; as a result, they have become the dominant paradigm in recent diffusion-based language modeling research. Unless otherwise specified, we use the term \emph{DLMs} in this paper to refer to discrete diffusion language models.

Discrete DLMs typically employ one of two attention mechanisms in the Transformer architecture: \emph{semi-causal attention} or \emph{full attention}. Models such as BD3-LM \cite{arriola2025block}, NBDiff \cite{tian2025next}, and Fast-dLLMv2 \cite{wu2025fast2} adopt semi-causal attention, which indicates that their predefined attention maps are incomplete and tokens attend only to their current "block" and previous ones. In contrast, models such as LLaDA \cite{nie2025large} and Dream \cite{ye2025dream} adopt full attention, allowing each token to condition on the entire sequence during decoding. In this work, we consider both semi-causal and full-attention DLMs to demonstrate the generalization and robustness of the proposed DCD decoding algorithm across different architectural choices.





\subsection{Decoding Strategies of DLMs}
Earlier works on DLMs \cite{austin2021structured,sahoo2024simple} randomly unmask and remask a fixed number of tokens at each decoding step, which often yields suboptimal performance. Later approaches incorporate confidence- or entropy-based criteria, decoding tokens whose uncertainty falls below a threshold or lies within the top-k candidates. These strategies improve flexibility and parallelism but still rely on fixed decoding ranges.

More recently, a variety of decoding strategies have been introduced to enhance either the performance or computational efficiency of discrete diffusion language models (DLMs). Among training-based approaches, \cite{xu2024energy} employs energy functions to steer the decoding process, yielding a 1.3x speedup alongside notable gains in generation quality. FS-DFM \cite{monsefi2025fs} formulates a discrete flow-matching framework that generates 1024 tokens in just eight sampling steps without degrading perplexity, while SDLM \cite{liu2025sequential} adaptively decodes token sequences based on prediction confidence.

For training-free methods, \cite{fu2025bits} introduces an Explore-Then-Exploit scheduling mechanism that maximizes information gain per decoding round to improve efficiency. \cite{chen2025beyond} enhances decoding quality by leveraging historical trajectories to inform current predictions, and \cite{li2025diffusion} proposes early-commit decoding to accelerate inference in DLMs while largely preserving output fidelity. Despite these advances, existing training-free strategies still lack mechanisms to dynamically adjust the decoding horizon or to enrich contextual support for low-certainty tokens, indicating significant opportunities for further innovation.

\section{Preliminary of DLMs Decoding}

\subsection{Formulations of DLMs}

Discrete diffusion language models (DLMs) generate a target sequence 
$\mathbf{x} = (x_1, \dots, x_T)$ by iteratively denoising a partially masked sequence. 
At diffusion step $t$, the sequence $\mathbf{x}^{(t)}$ contains masked positions denoted by 
$\langle \text{MASK} \rangle$. The reverse denoising process is modeled as:
\begin{equation}
  p_\theta(\mathbf{x}^{(t-1)} \mid \mathbf{x}^{(t)}) 
  = \prod_{i \in \mathcal{M}^{(t)}} p_\theta(x_i \mid \mathbf{x}^{(t)}),
\end{equation}
where $\mathcal{M}^{(t)}$ denotes the set of masked positions at step $t$.

For \emph{full-attention} DLMs, each masked token $x_i$ is predicted by conditioning on the entire partially decoded sequence $\mathbf{x}^{(t)}$. 
In contrast, \emph{semi-causal} DLMs partition the sequence into ordered blocks 
$\{\mathcal{B}_1, \dots, \mathcal{B}_K\}$ and restrict attention such that tokens in block $\mathcal{B}_k$ are conditioned only on tokens from blocks $\{\mathcal{B}_1, \dots, \mathcal{B}_k\}$. 
Accordingly, the reverse process can be written as:
\begin{equation}
p_\theta(\mathbf{x}^{(t-1)} \mid \mathbf{x}^{(t)}) 
= \prod_{k=1}^{K} \prod_{i \in \mathcal{B}_k \cap \mathcal{M}^{(t)}} 
p_\theta(x_i \mid \mathbf{x}^{(t)}_{\le k}),
\end{equation}
where $\mathbf{x}^{(t)}_{\le k}$ denotes the tokens in the first $k$ blocks.

\subsection{Block-based decoding of DLMs}

Decoding proceeds by selecting token values for a subset of masked positions according to the model prediction:
\begin{equation}
  \label{eq:pred}
  x_i^{(t-1)} =
  \begin{cases}
  \arg\max_{v \in \mathcal{V}} p_\theta(v \mid \mathbf{x}^{(t)}), & i \in \mathcal{S}^{(t)}, \\
  x_i^{(t)}, & \text{otherwise},
  \end{cases}
\end{equation}
where $\mathcal{V}$ is the vocabulary set and $\mathcal{S}^{(t)} \subseteq \mathcal{M}^{(t)}$ specifies the positions eligible for decoding at the current step.

In block-based decoding, the eligibility condition constrains decoding positions to a fixed region, typically the current block: $\mathcal{S}^{(t)} \subseteq \{i \mid i \in \mathcal{B}_{\text{cur}}\}$. Full-attention DLMs may optionally adopt block-based decoding to improve KV-cache compatibility. In contrast, semi-causal DLMs must employ block-based decoding due to their blockwise attention constraints. Within a large attention block, semi-causal DLMs may further apply sub-block decoding, where decoding positions are restricted to a smaller contiguous region: $\mathcal{S}^{(t)} \subseteq \{i \mid i \in \mathcal{B}_{\text{cur}_1:\text{cur}_2}\}$, where $\mathcal{B}_{\text{cur}_1:\text{cur}_2}$ denotes a contiguous subrange of blocks within a larger attention block.

\section{Boundary-Induced Context Truncation}

The advantages of DLMs over their autoregressive counterparts lie primarily in their (semi-)bidirectional attention horizons. \cite{piskorz2025masks} found that DLMs exhibit a strong contextual locality bias, in which nearby tokens $\mathbf{x}_{[i-\omega_l:i+\omega_r]}$ contribute disproportionately to prediction certainty of $i$-th token. Informally, this can be expressed as:
\begin{equation}
  \label{eq:a}
  p_\theta(x_i \mid \mathbf{x}^{(t)}) \approx p_\theta(x_i \mid \mathbf{x}^{(t)}_{[i-\omega_l:i+\omega_r]})
\end{equation}
However, under block-based decoding, tokens after the current block are all $\langle \text{MASK} \rangle$, which contain little information and may even distract the decoding process. For tokens whose contextual locality extends beyond the current block, Equation~\ref{eq:a} deteriorates to
\begin{equation}
  \label{eq:b}
  p_\theta(x_i \mid \mathbf{x}^{(t)}) \approx p_\theta(x_i \mid \mathbf{x}^{(t)}_{[i-\omega_l:b]})
\end{equation}
where $b < i + \omega_r$ denotes the right boundary of the current block. We term this reduction in a token's effective contextual window the \textbf{Boundary-Induced Context Truncation} phenomenon. Although these tokens receive insufficient context, they must be decoded before proceeding to the next block under the block-based paradigm. Consequently, this leads to low-certainty decoding at the end of each block and ultimately degrades the generation performance of DLMs.

\begin{figure}[h]
  \includegraphics[width=\columnwidth]{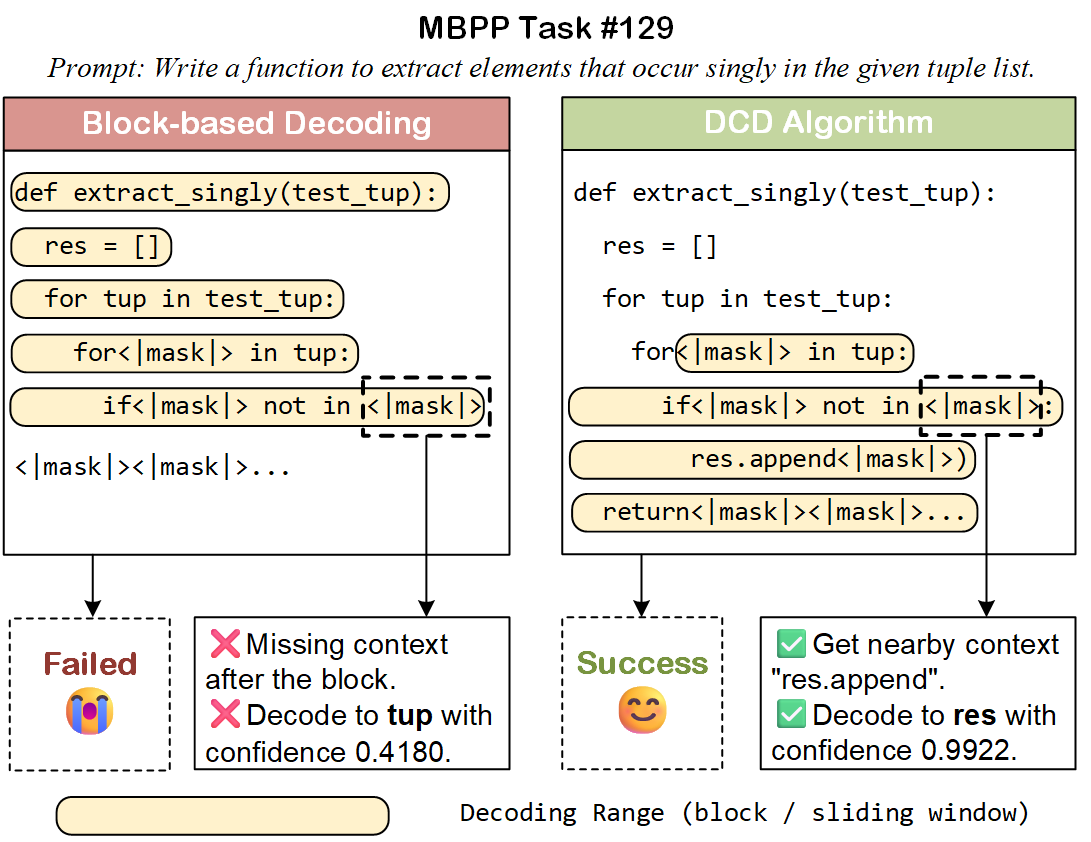}
  \caption{\textbf{An example of DCD and block-based decoding.} DCD outperforms block-based decoding by solving BICT problem. More details can be found in Appendix~B due to page limit.}
  \label{fig:case_study}
\end{figure}


\begin{figure*}[htbp]
  \centering
  \begin{minipage}[b]{0.32\textwidth}
      \centering
      \includegraphics[width=\linewidth]{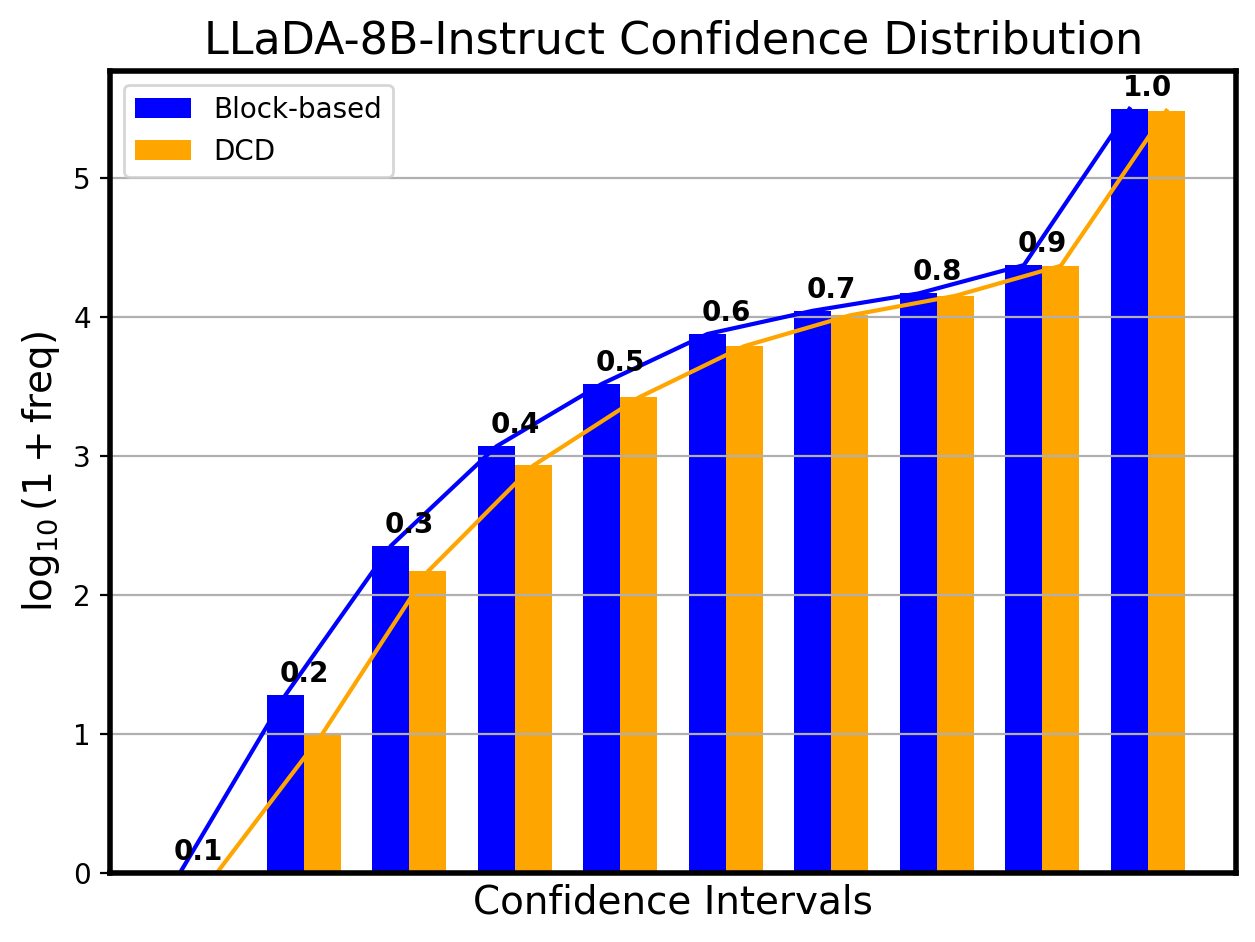}
  \end{minipage}
  \hfill
  \begin{minipage}[b]{0.32\textwidth}
      \centering
      \includegraphics[width=\linewidth]{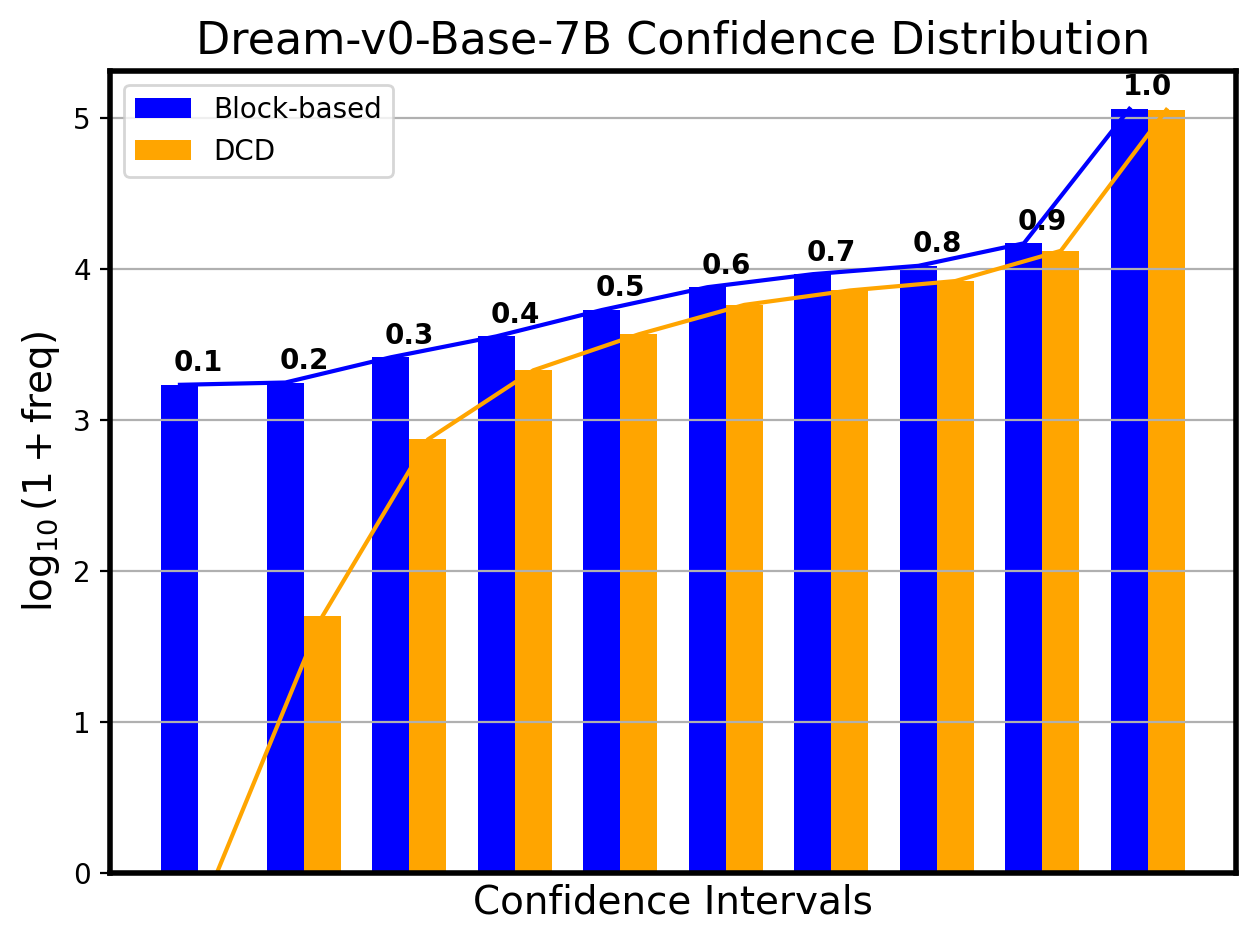}
  \end{minipage}
  \hfill
  \begin{minipage}[b]{0.32\textwidth}
      \centering
      \includegraphics[width=\linewidth]{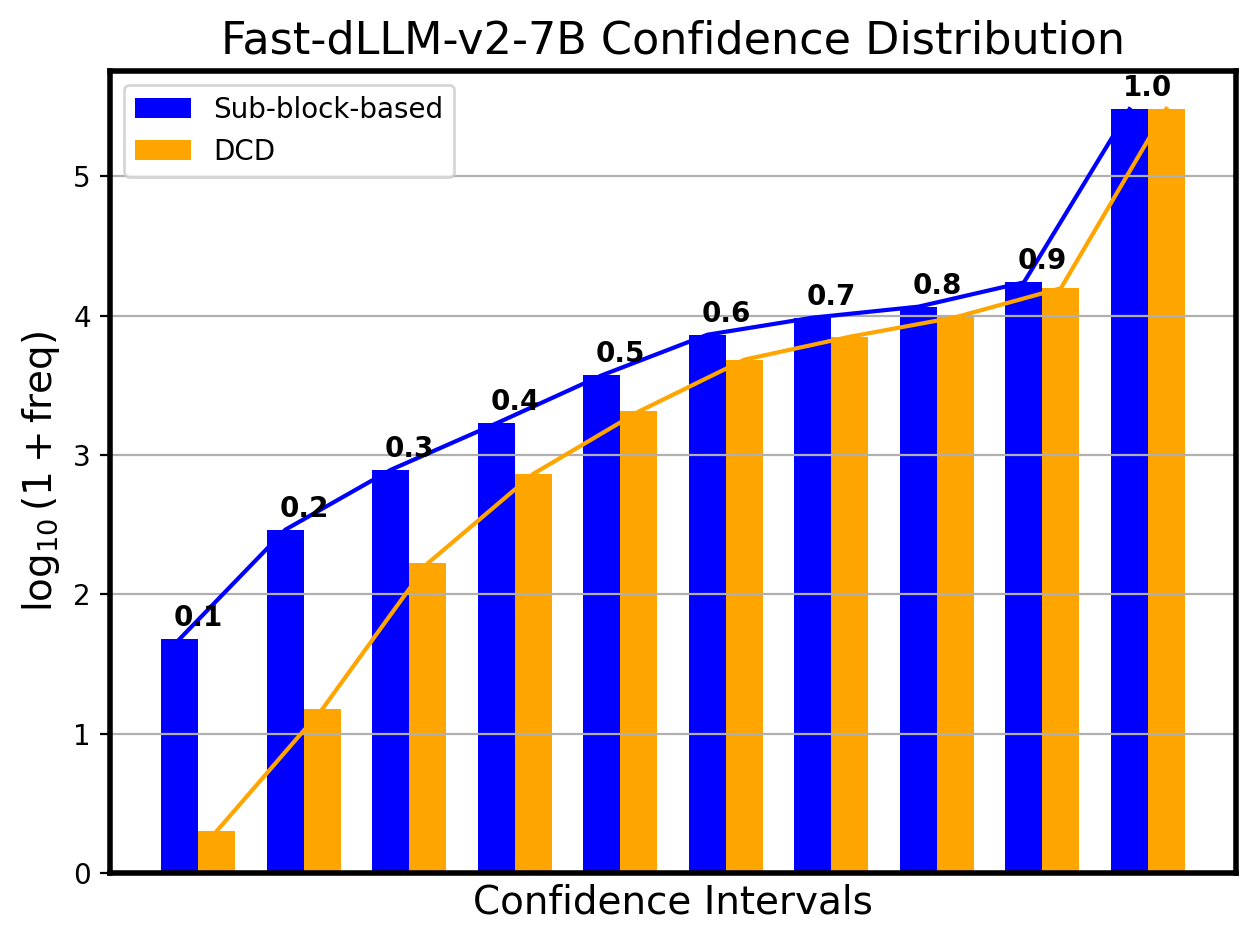}
  \end{minipage}
  \caption{\textbf{Confidence distributions of various models.} We use LLaDA-8B-Instruct, Dream-v0-Base-7B and Fast-dLLM-v2-7B on GSM8K with DCD and (sub-)block-based decoding using dual cache and log$_{10}$ scale for clarity. The DCD algorithm yields fewer low-certainty decoding steps across these models.}
  \label{fig:confidence}
\end{figure*}

\section{Deferred Commitment Decoding}

\subsection{Core Design of the DCD Algorithm}

Based on the above analysis, the primary causes of BICT are rigid block boundaries and the strict left-to-right blockwise decoding order. To address this problem, we must:
\begin{itemize}
  \item[(1)] Remove the restrictions imposed by rigid block boundaries;
  \item[(2)] Decode these tokens at the appropriate time and under appropriate conditions.
\end{itemize}
As illustrated in Figure~\ref{fig:overview}, the proposed \textbf{Deferred Commitment Decoding (DCD)} algorithm maintains a sliding window and defers the decoding of low-certainty tokens. It follows two principal design principles to achieve the above goals:

\textbf{Design (1): The decoding window slides left to right with constraints.} The sliding window defines the range of tokens eligible for decoding. It abandons fixed boundaries across consecutive decoding steps; instead, it moves from left to right within the \emph{generation slot} of the DLM. Formally, let $[L^{(t)}, R^{(t)})$ denote the left and right endpoints of the sliding window, and let $\mathbf{x}^{(t)}_{[l:r]}$ denote the generation slot at decoding step $t$. Then,
\begin{align}
  L^{(t)} &= \arg\min_{i \ge l} \{ i \mid x^{(t)}_{i} = \langle \text{MASK} \rangle \}, \label{eq:Lt} \\
  \begin{split}
    R^{(t)} &= \arg\max_{i \le r} \{ i \mid i \le L^{(t)} + s_{\text{max}} \text{ and } \\
    &\phantom{{}={}} \sum_{k=L^{(t)}}^{i-1} \left[ x^{(t)}_{k} = \langle \text{MASK} \rangle \right] \le s_{\text{init}} \}.
    \label{eq:Rt}
  \end{split}
\end{align}
Equation~\ref{eq:Lt} shows that the left endpoint of the window is anchored to the leftmost masked token. Equation~\ref{eq:Rt} indicates that the sliding window expands its right boundary as much as possible, subject to two constraints: (1) the total length of the sliding window does not exceed $s_{\text{max}}$, and (2) the number of masked tokens within the window does not exceed $s_{\text{init}}$. In particular, the sliding window is initialized with length $s_{\text{init}}$ at the beginning of the generation slot.

These constraints maintain a moderate yet flexible window length, which precisely captures relevant contextual information and enables more efficient KV-cache integration.

\textbf{Design (2): Tokens are deferred from decoding until sufficiently certain given nearby context.} Low prediction certainty—measured by the confidence or entropy of masked tokens—serves as a key indicator for identifying BICT-affected tokens with insufficient context. In block-based decoding, tokens may be forcibly decoded to complete a block. In contrast, the proposed DCD algorithm handles them more gracefully: masked tokens are decoded only when their certainty exceeds threshold $\tau$ or ranks among the top in the window.
\begin{equation}
  \label{eq:decode}
  \mathcal{S}^{(t)}=\mathcal{M}^{(t)} \cap [L^{(t)},R^{(t)}) \cap \{i \mid \mathcal{C}(i) \ge \min(\tau, \max_j \mathcal{C}(j)) \}
\end{equation}
where the certainty metric $\mathcal{C}(i)$ of the $i$-th token is computed using confidence or negative entropy. This approach significantly reduces low-certainty decoding events at later stages, thereby improving overall performance.

\emph{How does the DCD algorithm differ from AdaBlock?} The AdaBlock method \cite{lu2025adablock} employs adaptive block sizes based on delimiter semantics in the generated tokens, improving generation coherence. However, once determined, the block sizes remain fixed; as a result, AdaBlock may still suffer from BICT and thus leaves room for further improvement.

\begin{algorithm}[t]
  \caption{Deferred Commitment Decoding (DCD)}
  \label{alg:DCD}
  \begin{algorithmic}[1]
  \REQUIRE Generation slot $\mathbf{x}^{(t)}_{[l:r]}$, DLM $p_\theta(\cdot)$, window parameters $s_{\text{init}}, s_{\text{max}}$, cache parameters $\text{cache\_type}, B', r$, certainty threshold $\tau$, DBE parameters $\tau_{\text{low}},e_{\text{step}},e_{\text{max}}$.
  \STATE Initialize $L^{(t)}, R^{(t)} = l, l + s_{\text{init}}$, cache refresh $cd = 0$.
  \WHILE{$\mathcal{M}^{(t)} \neq \emptyset$}
    \IF{cache\_type $\ne$ none \AND $cd \le 0$}
      \STATE Refresh the cache based on Equation~\ref{eq:w}.
      \STATE Set $cd = B'$.
    \ENDIF
    \IF{Equation~\ref{eq:dbe1} is hold}
      \STATE Update $r,R^{(t-1)}$ based on Equation~\ref{eq:dbe2},\ref{eq:Rt}.
      \STATE \textbf{continue}
    \ENDIF
    \STATE Select decoding positions $\mathcal{S}^{(t)}$ using Equation~\ref{eq:decode}.
    \STATE Update $\mathbf{x}^{(t-1)}$ with $\mathcal{S}^{(t)}$ using Equation~\ref{eq:pred}.
    \STATE Update $L^{(t-1)}, R^{(t-1)}$ with $\mathbf{x}^{(t-1)}_{[l:r]}$ using Equations~\ref{eq:Lt} and~\ref{eq:Rt}.
    \STATE Update $cd = cd - |\mathcal{S}^{(t)}|$.
    \STATE Update $t = t - 1$.
  \ENDWHILE
  \RETURN Final sequence $\mathbf{x}^{(t)}_{[l:r]}$.
  \end{algorithmic}
\end{algorithm}

\subsection{Dynamic Block Extension for Semi-causal DLMs}

The core design of DCD fits well with full-attention DLMs because their generation slots span the entire sequence, and DCD completely replaces block-based decoding. However, for semi-causal DLMs, the large block size is predefined, and vanilla DCD operates only at the intra-block level by replacing fixed-length sub-block decoding with small sliding windows. Therefore, we propose \textbf{Dynamic Block Extension (DBE)}, a patch to DCD for semi-causal DLMs trained with multiple block sizes. Specifically, when a low-certainty token is about to be committed and the window's sliding is blocked by the right boundary of the large block—i.e., the BICT problem arises due to rigid block boundaries—the following condition holds:
\begin{equation}
  \label{eq:dbe1}
  \begin{split}
  \min_{i \in \mathcal{S}^{(t)}} \mathcal{C}(i) < \tau_{\text{low}} \quad \text{and} \quad R^{(t)} - L^{(t)} < s_{\text{max}} \\
  \text{and} \quad \sum_{k=L^{(t)}}^{R^{(t)}-1} \left[ x^{(t)}_{k} = \langle \text{MASK} \rangle \right] < s_{\text{init}}.
  \end{split}
\end{equation}
When this occurs, DBE aborts the current decoding step and expands the block with an upper limit:
\begin{equation}
  \label{eq:dbe2}
  r' = \min(r + e_{\text{step}},\ \text{blocksize} + e_{\text{max}}).
\end{equation}
Afterward, the sliding window is recalculated according to Equation~\ref{eq:Rt}, and a new decoding step begins to continue the generation loop. Although DBE relies on the DLM's ability to generalize to variable block sizes, this patch can further enhance performance for models appropriately trained for such flexibility.

\subsection{DCD's Combination with KV Cache}

To accelerate DLM inference, we integrate prefix and dual caching into the DCD algorithm, following Fast-dLLM \cite{wu2025fast}. Inspired by dKV-Cache-Greedy \cite{ma2025dkv}, the active interval without caching is slightly extended beyond the decoded tokens from the current and previous steps. Formally, it is defined as:
\begin{equation}
  \label{eq:w}
  \mathcal{W}^{(t)} = \left\{ x^{(t)}_i \mid i \in [L^{(t-1)} - r, R^{(t)} + r) \right\}.
\end{equation}
We then define the prefix of the generation slot as the tokens preceding $\mathcal{W}^{(t)}$, and the suffix as the tokens following $\mathcal{W}^{(t)}$. The prefix cache temporarily stores the prefix, while the dual cache stores both the prefix and suffix. Additionally, to ensure a fair comparison with block-based cache refreshing, we rebuild the cache after $B'$ masked tokens have been decoded since the last cache refresh.

\section{Experiments}  

\definecolor{darkgreen}{RGB}{0, 127, 0}
\begin{table*}[t]
  \centering
  \caption{\textbf{Experimental results.} For each experiment, we report its overall metrics (pass@1, accuracy, etc.). We also report the total seconds for running 5 benchmarks within one line. The best result of certain model and task is \textbf{bolded} and second-best is \underline{underlined}. The (sub-)block-based decoding serves as DCD's baseline and other algorithms serve as comparison works.}
  \label{tab:main}
  \resizebox{\textwidth}{!}{%
  \begin{tabular}{ccccccccc}
  \toprule
  
  \textbf{Model} & \textbf{Cache} & \textbf{Decoding} & \textbf{Time} & \textbf{Humaneval} & \textbf{MBPP} & \textbf{MATH500} & \textbf{GSM8K} & \textbf{IFEval}\\
  \midrule
  
  \multirow{10}{*}{%
  \begin{tabular}{c}
  LLaDA-8B \\
  -Instruct \\
  \textcolor{red}{Avg. Metric +1.16} \\
  \textcolor{darkgreen}{Avg. Time 0.0\%}
  \end{tabular}}
   & None & Block-based & 40750& 43.3 & 39.8 & 40.2 & 78.3 & \underline{57.9} \\
   & None & DCD & 40745& 43.9 & \textbf{40.0} & \underline{41.0} & \underline{79.1} & \textbf{59.0} \\
   & Prefix & Block-based & 24671& 43.3 & \underline{39.8} & 38.8 & 76.0 & 56.4 \\
   & Prefix & DCD & 24803& \textbf{45.7} & 38.2 & \textbf{41.2} & 78.5 & 57.1  \\
   & Dual & Block-based & 18617& 44.5 & 36.4 & 36.2 & 75.7 & 53.2 \\
   & Dual & DCD & 18501& 44.5 & 37.2 & 39.0 & \textbf{79.2} & 53.6  \\
  \cmidrule(lr){2-9}
   & \multicolumn{2}{c}{dKV-Cache-Greedy} & - & 15.37 & 20.4 & 27.0 & 68.23 & - \\
   & Dual & AdaBlock & 23680& \underline{45.1} & 36.2 & 36.6 & 78.4 & 55.8 \\
   & Custom & CCD & - & 38.41 & 39.20 & - & 75.30 & - \\
   & None & Prophet & - & 30.5 & 37.4 & - & 77.9 & - \\
  \midrule
  
  \multirow{8}{*}{
  \begin{tabular}{c}
  Dream-v0- \\
  Instruct-7B \\
  \textcolor{red}{Avg. Metric +2.63} \\
  \textcolor{darkgreen}{Avg. Time -2.2\%}
  \end{tabular}}
   & None & Block-based & 23685& 54.3 & 55.0 & \underline{44.8} & 76.6 & 50.5 \\
   & None & DCD & 23449& 53.7 & 56.8 & 43.8 & 78.2 & 55.6 \\
   & Prefix & Block-based & 15420& 56.7 & 53.6 & 43.4 & 77.6 & 51.8 \\
   & Prefix & DCD & 15044& \underline{58.5} & 57.4 & 43.4 & \underline{78.6} & \underline{56.4}  \\
   & Dual & Block-based & 9273& 56.7 & 52.8 & 44.4 & 74.8 & 47.7 \\
   & Dual & DCD & 9284& \textbf{59.8} & \textbf{58.8} & \textbf{45.2} & 77.3 & \textbf{56.7}  \\
   \cmidrule(lr){2-9}
   & Custom & CCD & - & 57.31 & \underline{58.00} & - & \textbf{82.51} & - \\
   & None & Prophet & - & 55.5 & 54.6 & - & 75.2 & - \\
   \midrule

  \multirow{7}{*}{
  \begin{tabular}{c}
  Dream-v0- \\
  base-7B \\
  \textcolor{red}{Avg. Metric +0.77} \\
  \textcolor{darkgreen}{Avg. Time -9.2\%}
  \end{tabular}}
  & None & Block-based & 25594& 48.2 & 13.8 & 12.0 & 75.5 & - \\
  & None & DCD & 22714& 50.6 & \textbf{17.0} & 12.8 & \textbf{76.0} & - \\
  & Prefix & Block-based & 14600& \underline{57.3} & 13.6 & 12.6 & 74.5 & - \\
  & Prefix & DCD & 13283& 53.0 & \underline{16.0} & 12.8 & 74.4 & - \\
  & Dual & Block-based & 10189& \textbf{57.3} & 13.4 & \textbf{13.2} & 73.8 & - \\
  & Dual & DCD & 9406& 56.1 & 13.2 & \underline{13.2} & 74.7 & - \\
  \cmidrule(lr){2-9}
  & Dual & AdaBlock & 42141 & 53.0 & 14.4 & 13.0 & \underline{76.0} & - \\
  \midrule

  \multirow{5}{*}{
  \begin{tabular}{c}
  Fast-dLLM \\
  -v2-7B \\
  \textcolor{red}{Avg. Metric +0.62} \\
  \textcolor{darkgreen}{Avg. Time -8.3\%}
  \end{tabular}}
   & None & Sub-block-based & 11228& \underline{61.0} & \textbf{50.2} & \textbf{54.6} & 77.6 & \underline{62.8} \\
   & None & DCD & 10116 & \textbf{62.8} & 48.6 & \underline{53.4} & \textbf{77.9} & \textbf{64.0} \\
   & Dual & Sub-block-based & 11379& 57.9 & 46.0 & 52.4 & 76.0 & 60.3  \\
   & Dual & DCD & 10498 & 59.1 & \underline{49.0} & 51.6 & \underline{77.8} & 60.8 \\
  \cmidrule(lr){2-9}
   & None & Block-based & 9993& 56.7 & 48.4 & 50.8 & 74.5 & 62.5 \\
  \midrule

  \multirow{2}{*}{
    \begin{tabular}{c}
    NBDiff \\
    \textcolor{red}{Avg. Metric +5.22} \\
    \end{tabular}}
     & None & Sub-block-based & - & \textbf{82.3} & \underline{78.2} & \underline{80.1} & \underline{87.4} & \underline{40.1} \\
     & None & DCD & 526163 & \underline{81.7} & \textbf{80.2} & \textbf{84.4} & \textbf{91.3} & \textbf{56.6} \\
   \bottomrule
  \end{tabular}
  }
\end{table*}

\begin{figure*}[htbp]
  \centering
  \begin{minipage}[b]{0.32\textwidth}
      \centering
      \includegraphics[width=\linewidth]{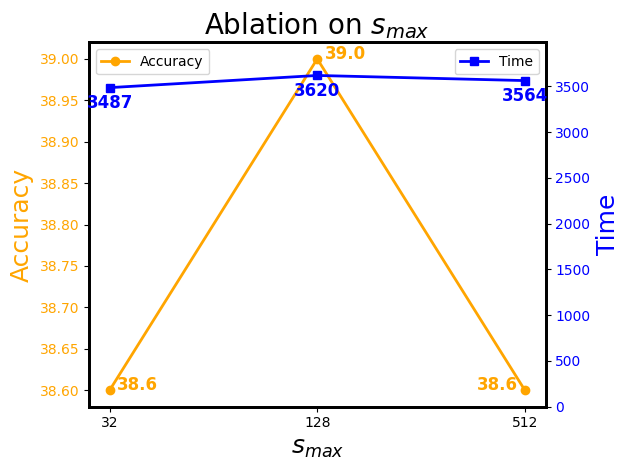}
  \end{minipage}
  \hfill
  \begin{minipage}[b]{0.32\textwidth}
      \centering
      \includegraphics[width=\linewidth]{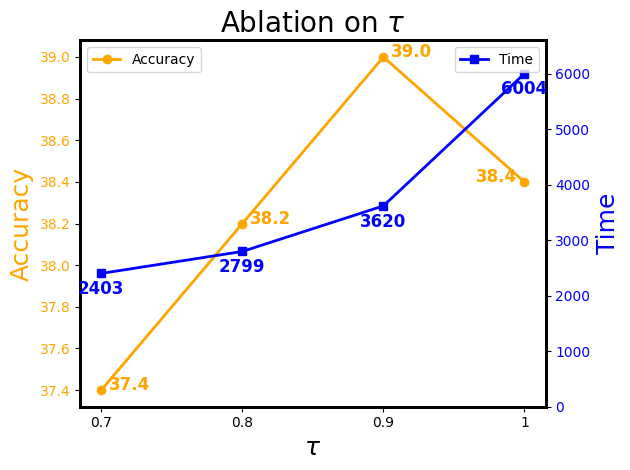}
  \end{minipage}
  \hfill
  \begin{minipage}[b]{0.32\textwidth}
      \centering
      \includegraphics[width=\linewidth]{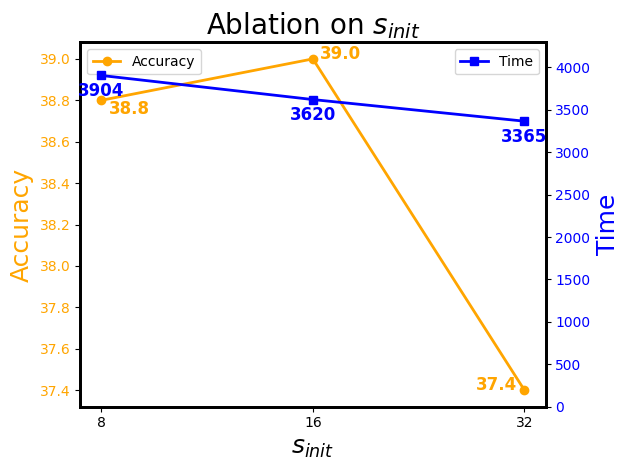}
  \end{minipage}
  \caption{\textbf{Ablation studies.} We vary $s_{\text{max}}$, $s_{\text{init}}$, and $\tau$, and evaluate task accuracy and decoding time on LLaDA-8B-Instruct on the MATH500 task with dual-cache DCD.}
  \label{fig:ablation_study}
\end{figure*}

\subsection{Experimental Setup}

\paragraph{Models.}
To validate the effectiveness of the DCD algorithm, we evaluate full-attention pretrained DLMs including LLaDA-8B-Instruct \cite{nie2025large}, Dream-v0-Instruct-7B, and Dream-v0-Base-7B \cite{ye2025dream}; and semi-causal DLMs including Fast-dLLM-v2-7B \cite{wu2025fast2} and NBDiff \cite{tian2025next}.

\paragraph{Benchmarks.}
For each model, we evaluate coding benchmarks including HumanEval \cite{humaneval} and MBPP \cite{mbpp}, mathematical reasoning benchmarks including MATH500 \cite{math500} and GSM8K \cite{gsm8k}, and the instruction-following benchmark IFEval \cite{ifeval}. The Dream-v0-Base-7B model is excluded from IFEval because it is not instruction-aligned. We do not evaluate multiple-choice QA benchmarks \cite{mmlu,gpqa}, as they primarily measure token-level log-probabilities rather than decoding quality.

\paragraph{Cache Configurations and Baselines.}
For full-attention DLMs, the parallel block-based decoding of Fast-dLLM \cite{wu2025fast} serves as the baseline under three cache configurations: no cache, prefix cache, and dual cache. For the semi-causal DLM Fast-dLLM-v2-7B \cite{wu2025fast2}, sub-block-based decoding with no cache and with dual cache within each large block is used as the primary baseline, while vanilla block-based decoding without sub-block structures is used as an additional baseline. For each model, benchmark, and cache configuration, DCD is compared against these baselines as well as other training-free DLM decoding strategies such as dKV-Cache \cite{ma2025dkv}, AdaBlock-dLLM \cite{lu2025adablock}, CCD \cite{chen2025beyond}, and Prophet \cite{li2025diffusion} when available.

\paragraph{Hyperparameters.}
All experiments use block size $B=32$, sub-block size $b=8$ and $\text{batchsize}=1$. To align with the configurations of baseline works, decoding certainty is measured by negative entropy for NBDiff \cite{tian2025next} and by confidence for the remaining DLMs \cite{wu2025fast,wu2025fast2}, with a unified threshold $\tau=0.9$. For full-attention DLMs, we set $L=512$, $s_{\text{init}}=16$, $s_{\text{max}}=128$, $B'=32$ and $r=2$. For the semi-causal DLM, we set $s_{\text{init}}=8$ while keeping all other parameters unchanged. The DBE is not enabled in main results and discussed in Section~\ref{sec:dbe} with $e_{\text{step}}=4,e_{\text{max}}=16$ and $\tau_{\text{low}}=0.4$ for confidence and $\tau_{\text{low}}=-0.5$ for negative entropy.

\subsection{Main Results Analysis}

Table~\ref{tab:main} reports the main experimental results across multiple diffusion language models, tasks, and KV-cache configurations. Overall, Deferred Commitment Decoding (DCD) consistently outperforms block-based and sub-block-based decoding across most settings, demonstrating strong robustness across tasks and model architectures. On average, DCD improves evaluation metrics by \textbf{+1.73\%} while reducing decoding time by \textbf{4.4\%} compared to block-based (or sub-block-based) baselines with regard to the same model, task and cache configuration. The average improvement for each model over the baseline is highlighted in colored text in Table~\ref{tab:main}.

DCD's gains over (sub-)block-based baselines are observed across diverse tasks—including mathematical reasoning, code generation, and instruction following—and across both full-attention and semi-causal DLMs. Among all models, LLaDA-8B-Instruct, Dream-v0-Instruct-7B, and Dream-v0-Base-7B exhibit significant improvements in accuracy and moderate speedups in time consumption, showing that our DCD algorithm works effectively at the whole-sequence level. \textbf{NBDiff} benefits the most from DCD, achieving the largest average improvement of \textbf{5.22\%}, as it is trained with multiple block sizes and thus aligns well with sliding windows. Running the IFEval benchmark with NBDiff yields the most significant improvement, with a \textbf{16.5\%} increase in accuracy. In contrast, \textbf{Fast-dLLM-v2-7B} shows the smallest improvement of 0.62\% because it is trained with limited flexibility with variable decoding ranges. Nevertheless, DCD still outperforms sub-block-based decoding, which aligns with our theoretical analysis.

Compared with AdaBlock \cite{lu2025adablock}, DCD achieves better performance (average metric improvements of +0.28\% for LLaDA-8B-Instruct and +1.20\% for Dream-v0-Base-7B) and substantial speedups (average decoding time reductions of 19\% and 71\%, respectively), demonstrating the effectiveness of the deferred commitment mechanism. Compared with dKV-Cache-Greedy \cite{ma2025dkv}, DCD substantially outperforms it in terms of accuracy, despite employing a similar KV-cache strategy. Compared with CCD \cite{chen2025beyond}, which uses a custom cache to store historical data, the best results of DCD surpass it by 1.90\% on average. More importantly, DCD outperforms Prophet \cite{li2025diffusion} by a large margin because the latter algorithm contradicts DCD by performing early decoding, which substantially degrades DLMs' inference performance.

We note that in a small number of cases, DCD performs slightly worse than (sub-)block-based decoding and other methods. We attribute these regressions to the inherent stochasticity of training-free decoding and the intrinsic difficulty of certain tokens, which may remain ambiguous even with extended context.

\subsection{Evidence of BICT Mitigation}

Figure~\ref{fig:confidence} provides direct evidence that DCD mitigates Boundary-Induced Context Truncation. We visualize the distribution of decoding confidence on the GSM8K benchmark for LLaDA-8B-Instruct, Dream-v0-Base-7B, and Fast-dLLM-v2-7B. GSM8K is selected because it is the largest benchmark and exhibits the most stable improvements under DCD.

Across all models, DCD substantially reduces the frequency of extremely low-certainty decoding steps compared to block-based or sub-block-based decoding. Such low-certainty events directly reflect the BICT phenomenon that DCD is designed to address. This reduction provides a clear explanation for the observed accuracy improvements, particularly on reasoning-intensive tasks.

\subsection{Ablation Studies}

Figure~\ref{fig:ablation_study} presents ablation studies on LLaDA-8B-Instruct evaluated on MATH500 with dual cache enabled, analyzing the impact of three key hyperparameters in DCD: the maximum window size $s_{\max}$, the initial window size $s_{\text{init}}$, and the certainty threshold $\tau$.

\paragraph{Effect of $s_{\max}$.}
The maximum window size determines the upper bound of contextual expansion. When $s_{\max}=32$, the window is constrained to the baseline block size, limiting DCD's ability to mitigate BICT. When $s_{\max}=512$ (equivalent to removing the upper bound given $L=512$), accuracy degrades due to diluted contextual relevance. Decoding time does not exhibit a consistent monotonic trend with respect to $s_{\max}$, as the window rarely expands to its maximum in practice.

\paragraph{Effect of $s_{\text{init}}$.}
The initial window size controls early decoding behavior. Setting $s_{\text{init}}=8$ reduces the available context and may degrade performance, while $s_{\text{init}}=32$ may lead to excessively long windows, introducing premature commitments near the right boundary. The weak negative correlation between $s_{\text{init}}$ and decoding time may result from increased parallelism enabled by larger windows.

\paragraph{Effect of $\tau$.}
The certainty threshold regulates how aggressively tokens are deferred. Accuracy improves as $\tau$ increases from lower values but degrades when $\tau=1$ (i.e., top-1 confidence decoding), as this setting becomes fully deterministic and loses flexibility. In contrast to window parameters, $\tau$ exhibits a clearer positive correlation with decoding time.

Overall, accuracy exhibits a clear unimodal trend as $s_{\max}$, $s_{\text{init}}$, and $\tau$ increase in this setting. Based on these ablation results, we select appropriate hyperparameters and apply them consistently across all experiments.

\subsection{The Effects of Dynamic Block Extention}
\label{sec:dbe}

Table~\ref{tab:dbe} reports the performance of DBE-patched DCD across five benchmarks for semi-causal DLMs. Despite minimal computational overhead, NBDiff consistently benefits from DBE across various tasks, achieving an average improvement of 1.14\%, as expected. This is because NBDiff's training data includes generation slots of varying block sizes. In contrast, Fast-dLLM-v2 is trained with a fixed block size of $B=32$, and thus fails to benefit from DBE, instead suffering a 1.30\% drop in metrics. Overall, we suggest that semi-causal DLMs should be trained with variable block sizes to mitigate the BICT problem and enable the DBE mechanism.

\begin{table}[ht]
  \centering
  \caption{\textbf{Experimental results for DBE-patched DCD.} The comparison baselines are vanilla DCD without cache for two models.}
  \label{tab:dbe}
  \begin{tabular}{c|c|c}
  \hline
  \textbf{Item} & \textbf{NBDiff} & \textbf{Fast-dLLM-v2} \\
  \hline
  HumanEval & $83.5^{\textcolor{red}{+1.8}}$ & $58.5^{\textcolor{darkgreen}{-4.3}}$ \\
  MBPP & $80.9^{\textcolor{red}{+0.7}}$ & $47.2^{\textcolor{darkgreen}{-1.4}}$ \\
  MATH500 & $86.2^{\textcolor{red}{+1.8}}$ & $54.0^{\textcolor{red}{+0.6}}$ \\
  GSM8K & $90.9^{\textcolor{darkgreen}{-0.4}}$ & $78.4^{\textcolor{red}{+0.5}}$ \\
  IFEval & $58.4^{\textcolor{red}{+1.8}}$ & $62.1^{\textcolor{darkgreen}{-1.9}}$ \\
  \hline
  Avg. Metric & $79.98^{\textcolor{red}{+1.14}}$ & $60.04^{\textcolor{darkgreen}{-1.30}}$  \\
  Time & $604526^{\textcolor{red}{+14.9\%}}$ & $10757^{\textcolor{red}{+6.3\%}}$ \\
  \hline
  \end{tabular}
\end{table}

\section{Conclusion}

In this work, we investigate a fundamental limitation of block-based diffusion decoding for language models, which we formalize as \textbf{Boundary-Induced Context Truncation}. We identify suboptimal token commitment—whereby tokens that would otherwise benefit from nearby future context are forced to commit at block boundaries—leading to low-certainty predictions and degraded generation quality. To address this issue, we propose \textbf{Deferred Commitment Decoding}, a simple, training-free decoding strategy that replaces fixed block boundaries with sliding windows. By deferring uncertain tokens until sufficient context becomes available, DCD enables more effective utilization of local bidirectional context without sacrificing KV-cache compatibility. Extensive experiments across multiple DLMs, benchmarks, and caching configurations demonstrate that DCD consistently improves generation accuracy by 1.73\% on average with comparable decoding time relative to fixed block-based baselines, with the maximum improvement reaching 16.5\%. These results demonstrate the superiority of DCD for DLM inference.



\bibliographystyle{named}
\bibliography{ijcai26}

\clearpage
\appendix
\renewcommand\thesection{\Alph{section}}
\definecolor{dkgreen}{rgb}{0,0.6,0}
\definecolor{gray}{rgb}{0.5,0.5,0.5}
\definecolor{mauve}{rgb}{0.58,0,0.82}
\lstset{
language=Python,
frame=tb,
aboveskip=3mm,
belowskip=3mm,
showstringspaces=false,
columns=flexible,
basicstyle={\small\ttfamily},
numbers=left,
numberstyle=\tiny\color{gray},
keywordstyle=\color{blue},
commentstyle=\color{dkgreen},
stringstyle=\color{mauve},
breaklines=true,
breakatwhitespace=true,
tabsize=4
}

\section{Experimental Environment}

\paragraph{Hardware.}
For comparison experiments between (sub-)block-based decoding and DCD on LLaDA-8B-Instruct, Dream-v0-Base-7B, and Fast-dLLM-v2-7B, we allocate one NVIDIA A100 80GB GPU and eight Intel(R) Xeon(R) Platinum 8358 CPUs for each experiment. For AdaBlock and NBDiff experiments, we allocate one NVIDIA A800 GPU and sixteen Intel(R) Xeon(R) Platinum 8378A CPUs for each experiment.

\paragraph{Software.}
We use Ubuntu Linux, CUDA 12, and Python 3.10 for all experiments. For LLaDA-8B-Instruct, Dream-v0-Base-7B, and Fast-dLLM-v2-7B, we follow their original papers and use \texttt{lm-eval-harness} as the evaluation suite. For NBDiff, we use \texttt{opencompass} as the evaluation suite.

\section{More Details about BICT Case Study}

Figure~2 in main paper illustrates the BICT phenomenon in block-based decoding. This example corresponds to the 129th test case of the MBPP benchmark, generated by Dream-v0-Instruct-7B without any caching. The prompt for this task is: ``Write a function to extract elements that occur singly in the given tuple list.'' The test cases are:
\begin{lstlisting}
assert extract_singly([(3, 4, 5), (4, 5, 7), (1, 4)]) == [3, 4, 5, 7, 1]
assert extract_singly([(1, 2, 3), (4, 2, 3), (7, 8)]) == [1, 2, 3, 4, 7, 8]
assert extract_singly([(7, 8, 9), (10, 11, 12), (10, 11)]) == [7, 8, 9, 10, 11, 12]
\end{lstlisting}

The incorrect code generated by the block-based method is:
\begin{lstlisting}
def extract_singly(test_tup):
  res = []
  for tup in test_tup:
    for i in tup:
      if i not in tup:
        res.append(i)
  return res
\end{lstlisting}

The error occurs at decoding step 21, as illustrated in the left part of Figure~2. The $\langle \text{MASK} \rangle$ at a critical position (the rightmost token of line 5) is adjacent to the first block boundary and suffers from truncated context. As a result, it is incorrectly decoded as "tup" with low confidence.

The correct code generated by the DCD method is:
\begin{lstlisting}
def extract_singly(test_tup):
  res = []
  for tup in test_tup:
    for num in tup:
      if num not in res:
        res.append(num)
  return res
\end{lstlisting}

The critical decoding step 23 is illustrated in the right part of Figure~2. The DCD algorithm successfully resolves this issue by deferring the decoding of the critical $\langle \text{MASK} \rangle$ until the sliding window incorporates sufficient contextual information, such as ``res.append''. Consequently, the model correctly decodes the second occurrence of ``res'' and successfully passes all test cases.

\section{Details about Main Experiments}

\subsection{lm\_eval\_harness Part}

For LLaDA-8B-Instruct, Dream-v0-Base-7B and Fast-dLLM-v2-7B, we use lm-eval-harness 0.4.8 and the code-cleaning suite in the Fast-dLLM codebase. Specifically:
\begin{itemize}
  \item For \textbf{HumanEval}, we use 0-shot pass@1 as the evaluation metric. For code cleaning, we concatenate the prompt and the generated output, remove possible ```python ... ``` blocks, and extract the function based on Python syntax trees. No prompt engineering is applied in this setting.
  \item For \textbf{MBPP}, we use 3-shot pass@1 as the evaluation metric. The code-cleaning logic is the same as that for HumanEval. Prompt engineering and the few-shot mechanism are automatically handled by the lm-eval-harness library.
  \item For \textbf{MATH500}, we use 0-shot accuracy as the evaluation metric, with simple chain-of-thought reasoning prompts as the default. The cleaning logic extracts the boxed answer and simplifies the mathematical expression.

\lstset{
  basicstyle=\ttfamily,
  columns=flexible,
  keepspaces=true,
  showstringspaces=false,
  breaklines=true,
  frame=tb,
  numbers=none,
  escapechar=,
  literate=,
  language=,
  keywordstyle=,
  commentstyle=,
  stringstyle=,
  basicstyle=\small\ttfamily
}

  \begin{lstlisting}
You are a math expert. You will be given a question to solve. Solve it step by step. Wrap the final answer in a \\boxed{}.
Respond in the following format:
<reasoning>
Your reasoning here
</reasoning>
<answer>
boxed{...}
</answer>
{{text}}
  \end{lstlisting}
  \item For \textbf{GSM8K}, we use 5-shot accuracy as the evaluation metric, corresponding to the ``exact\_match,flexible-extract'' value reported in the lm-eval-harness result files. All other settings follow the default configuration.
  \item For \textbf{IFEval}, we use 0-shot accuracy as the evaluation metric, corresponding to the ``prompt\_level\_strict\_acc,none'' value reported in the lm-eval-harness result files. All other settings follow the default configuration.
\end{itemize}
Notably, we do not use any stop words (e.g., ``[DONE]'' in the default MBPP configuration) in any experiments. This choice may lead to degraded performance for Dream-v0-Base-7B on these tasks, as unaligned models often struggle to terminate generation appropriately and may produce extraneous content after completing the task. However, this setting is applied consistently across all experiments in this paper, ensuring fair comparisons.

\lstset{
  basicstyle=\ttfamily,
  columns=flexible,
  keepspaces=true,
  showstringspaces=false,
  breaklines=true,
  frame=tb,
  numbers=none,
  escapechar=,
  literate=,
  language=,
  keywordstyle=,
  commentstyle=,
  stringstyle=,
  basicstyle=\small\ttfamily
}

\subsection{OpenCompass Part}

For NBDiff, we use zero-shot evaluation with simple or no prompt engineering, as detailed below. The scoring logic is entirely based on \url{https://github.com/open-compass/opencompass}. These configurations are identical to those in the original paper.

\begin{itemize}
  \item \textbf{HumanEval:}
  \begin{lstlisting}
    Complete the following python code:
    {{text}}
  \end{lstlisting}
  \item \textbf{MBPP:}
  \begin{lstlisting}
    You are an expert Python programmer, and here is your task: {{text}}
    Your code should pass these tests:
    
    {{test_list[0]}}
    {{test_list[1]}}
    {{test_list[2]}}
    You should submit your final solution in the following format: ```python
    
    ```
  \end{lstlisting}
  \item \textbf{MATH500:}
  \begin{lstlisting}
    {{text}}
    Please reason step by step, and put your final answer within \\boxed{}.
  \end{lstlisting}
  \item \textbf{GSM8K:}
  \begin{lstlisting}
    Question: {{text}}
    Please reason step by step, and put your final answer within \\boxed{}.
  \end{lstlisting}
  \item \textbf{IFEval:} No additional prompt.
\end{itemize}

\section{More Statistics of Each Experiment}

\subsection{Time Consumption}

To better understand the DCD method, we record the time consumption for each benchmark experiment. According to Table~\ref{tab:time}, DCD completes benchmarks in comparable—and even slightly less—time than the baseline method, demonstrating its efficiency relative to traditional approaches. For dKV-Cache-Greedy, CCD, and Prophet, we copy their statistics from the original papers, as their time consumption data are unavailable.

\begin{table*}[t]
  \centering
  \caption{Detailed time consumption for each experiment. (unit: seconds)}
  \label{tab:time}
  \resizebox{\textwidth}{!}{%
  \begin{tabular}{cccccccc}
  \toprule
  
  \textbf{Model} & \textbf{Cache} & \textbf{Decoding} & \textbf{Humaneval} & \textbf{MBPP} & \textbf{MATH500} & \textbf{GSM8K} & \textbf{IFEval} \\
  \midrule
  
  \multirow{7}{*}{%
  \begin{tabular}{c}
  LLaDA-8B \\
  -Instruct
  \end{tabular}}
   & None & Block-based & 2128 & 7333 & 5958 & 18144 & 7187 \\
   & None & DCD & 2027 & 7467 & 5702 & 18167 & 7382 \\
   & Prefix & Block-based & 1536 & 3778 & 4551 & 8584 & 6222 \\
   & Prefix & DCD & 1565 & 3824 & 4478 & 8552 & 6384  \\
   & Dual & Block-based & \underline{1289} & \underline{2934} & \textbf{3594} & \underline{6356} & \textbf{4444} \\
   & Dual & DCD & \textbf{1252} & \textbf{2877} & \underline{3620} & \textbf{6266} & \underline{4486}  \\
  \cmidrule(lr){2-8}
   & Dual & AdaBlock & 1531 & 3934 & 3983 & 9350 & 4882 \\
  \midrule
  
  \multirow{6}{*}{
  \begin{tabular}{c}
  Dream-v0- \\
  Instruct-7B
  \end{tabular}}
   & None & Block-based & 1038 & 1688 & 5106 & 10186 & 5667 \\
   & None & DCD & 986 & 1588 & 5000 & 10299 & 5576 \\
   & Prefix & Block-based & 821 & 1022 & 3912 & 4922 & 4743 \\
   & Prefix & DCD & 783 & 948 & 3790 & 4844 & 4679  \\
   & Dual & Block-based & \underline{477} & \underline{587} & \underline{2587} & \textbf{2807} & \textbf{2815} \\
   & Dual & DCD & \textbf{470} & \textbf{586} & \textbf{2532} & \underline{2879} & \underline{2817}  \\
  \midrule

  \multirow{7}{*}{
  \begin{tabular}{c}
  Dream-v0- \\
  base-7B
  \end{tabular}}
  & None & Block-based & 1534 & 6777 & 5602 & 11681 & - \\
  & None & DCD & 1209 & 6987 & 5231 & 9287 & - \\
  & Prefix & Block-based & 1177 & 3963 & 3943 & 5517 & - \\
  & Prefix & DCD & 1066 & 4053 & 3729 & 4435 & - \\
  & Dual & Block-based & \underline{717} & \underline{2684} & \textbf{3594} & \underline{3194} & - \\
  & Dual & DCD & \textbf{658} & \textbf{2415} & \underline{3620} & \textbf{2713} & - \\
  \cmidrule(lr){2-8}
  & Dual & AdaBlock & 2243 & 7105 & 6432 & 26361 & - \\
  \midrule

  \multirow{5}{*}{
  \begin{tabular}{c}
  Fast-dLLM \\
  -v2-7B
  \end{tabular}}
   & None & Sub-block-based & 615 & 1566 & 2650 & 3285 & 3112 \\
   & None & DCD & \textbf{561} & \underline{1519} & \underline{2253} & \textbf{2793} & \underline{2990} \\
   & Dual & Sub-block-based & 670 & 1667 & 2428 & 3451 & 3163  \\
   & Dual & DCD & 598 & 1608 & 2339 & 2895 & 3058 \\
  \cmidrule(lr){2-8}
   & None & Block-based & \underline{561} & \textbf{1487} & \textbf{2168} & \underline{2831} & \textbf{2946} \\
 \midrule
 \multirow{1}{*}{
  \begin{tabular}{c}
  NBDiff
  \end{tabular}}
   & None & DCD & 35215 & 81001 & 179530 & 135040 & 95467 \\
 \bottomrule
  \end{tabular}
  }
\end{table*}

\subsection{Low-Certainty Decoding}

To better understand the BICT phenomenon, we collect the number of low-certainty ($\text{confidence} < 0.3$) decoding steps in each experiment. The results show that the DCD algorithm mitigates the BICT phenomenon by significantly reducing the number of low-certainty decoding steps, thereby improving performance.

\begin{table*}[t]
  \centering
  \caption{Low-certainty decoding steps of each experiment.}
  \label{tab:low}
  \resizebox{\textwidth}{!}{%
  \begin{tabular}{cccccccc}
  \toprule
  
  \textbf{Model} & \textbf{Cache} & \textbf{Decoding} & \textbf{Humaneval} & \textbf{MBPP} & \textbf{MATH500} & \textbf{GSM8K} & \textbf{IFEval} \\
  \midrule
  
  \multirow{6}{*}{%
  \begin{tabular}{c}
  LLaDA-8B \\
  -Instruct
  \end{tabular}}
   & None & Block-based & 57 & \underline{214} & 366 & 161 & 7450 \\
   & None & DCD & \underline{52} & \textbf{177} & \textbf{284} & \textbf{129} & \textbf{6848} \\
   & Prefix & Block-based & 67 & 220 & 370 & 184 & 7870 \\
   & Prefix & DCD & \textbf{50} & 269 & \underline{330} & \underline{137} & \underline{6975} \\
   & Dual & Block-based & 88 & 333 & 538 & 242 & 8181 \\
   & Dual & DCD & 82 & 245 & 421 & 155 & 7245 \\
  \midrule
  
  \multirow{6}{*}{
  \begin{tabular}{c}
  Dream-v0- \\
  Instruct-7B
  \end{tabular}}
   & None & Block-based & 181 & \textbf{43} & 473 & \underline{588} & \underline{7481} \\
   & None & DCD & \underline{165} & \underline{45} & \textbf{395} & \textbf{579} & \textbf{7381} \\
   & Prefix & Block-based & 193 & 66 & 508 & 647 & 7935 \\
   & Prefix & DCD & \textbf{159} & 54 & \underline{446} & 640 & 7493 \\
   & Dual & Block-based & 227 & 75 & 689 & 746 & 8304 \\
   & Dual & DCD & 196 & 59 & 565 & 702 & 7914 \\
  \midrule

  \multirow{6}{*}{
  \begin{tabular}{c}
  Dream-v0- \\
  base-7B
  \end{tabular}}
  & None & Block-based & 1393 & 951 & 1213 & 9111 & - \\
  & None & DCD & \textbf{353} & \textbf{631} & \textbf{888} & \textbf{601} & - \\
  & Prefix & Block-based & 1414 & 1129 & 1408 & 8726 & - \\
  & Prefix & DCD & \underline{399} & \underline{878} & \underline{899} & \underline{683} & - \\
  & Dual & Block-based & 1499 & 1277 & 1477 & 6084 & - \\
  & Dual & DCD & 471 & 971 & 894 & 791 & - \\
  \midrule

  \multirow{5}{*}{
  \begin{tabular}{c}
  Fast-dLLM \\
  -v2-7B
  \end{tabular}}
   & None & Sub-block-based & 244 & 701 & 292 & 805 & 5679 \\
   & None & DCD & \underline{184} & \underline{604} & \textbf{158} & \textbf{160} & \textbf{4512} \\
   & Dual & Sub-block-based & 332 & 994 & 401 & 1119 & 6748 \\
   & Dual & DCD & \textbf{174} & 690 & \underline{175} & \underline{181} & \underline{4751} \\
  \cmidrule(lr){2-8}
   & None & Block-based & 192 & \textbf{602} & 245 & 583 & 4861 \\
 \bottomrule
  \end{tabular}
  }
\end{table*}

\subsection{Other Efficiency Metrics}

The wall-clock time reported earlier is highly dependent on hardware-specific environments and exhibits limited generalization. Therefore, we report two additional metrics to assess the algorithmic efficiency of each experiment:

\begin{itemize}
  \item \textbf{Average decoding steps.} This metric measures the average number of decoding steps required to complete generation for prompts. It corresponds directly to the average number of forward propagations and constitutes the primary source of time consumption in DLM inference.
  \item \textbf{Average forward length.} This metric measures the average number of tokens directly fed into the DLM for forward propagation. In some scenarios, certain tokens are cached and thus excluded from this count. Consequently, this metric is positively correlated with both time consumption and memory usage per decoding step.
\end{itemize}

As shown in Table~\ref{tab:ds} and Table~\ref{tab:fl}, the average decoding steps of DCD are slightly lower (-5.0\% on average) than those of the (sub-)block-based baselines, indicating a modest efficiency gain from its decoding strategy. Meanwhile, the average forward length under DCD is comparable (+6.0\% on average) to that of the baselines across all models and tasks. Together, these results suggest that DCD incurs no additional computational overhead—in fact, its total inference time is on par with or slightly less than that of the (sub-)block-based approaches, consistent with the wall-clock time measurements.

\begin{table*}[t]
  \centering
  \caption{Average decoding steps of each experiment.}
  \label{tab:ds}
  \resizebox{\textwidth}{!}{%
  \begin{tabular}{cccccccc}
  \toprule
  
  \textbf{Model} & \textbf{Cache} & \textbf{Decoding} & \textbf{Humaneval} & \textbf{MBPP} & \textbf{MATH500} & \textbf{GSM8K} & \textbf{IFEval} \\
  \midrule
  
  \multirow{6}{*}{%
  \begin{tabular}{c}
  LLaDA-8B \\
  -Instruct
  \end{tabular}}
   & None & Block-based & 161.4 & 108.1 & 154.4 & 86.0 & 197.1 \\
   & None & DCD & 160.6 & 108.1 & 149.2 & 85.0 & 201.2 \\
   & Prefix & Block-based & 163.4 & 112.5 & 157.3 & 89.8 & 197.4 \\
   & Prefix & DCD & 165.2 & 113.4 & 151.5 & 87.0 & 199.5 \\
   & Dual & Block-based & 174.9 & 118.5 & 168.4 & 95.0 & 195.9 \\
   & Dual & DCD & 169.5 & 113.3 & 162.1 & 90.4 & 196.8 \\
  \midrule
  
  \multirow{6}{*}{
  \begin{tabular}{c}
  Dream-v0- \\
  Instruct-7B
  \end{tabular}}
   & None & Block-based & 88.5 & 28.0 & 150.1 & 55.7 & 176.7 \\
   & None & DCD & 85.9 & 26.4 & 147.9 & 55.0 & 179.7 \\
   & Prefix & Block-based & 88.1 & 28.8 & 152.0 & 56.5 & 172.5 \\
   & Prefix & DCD & 85.4 & 26.6 & 149.5 & 55.6 & 169.2 \\
   & Dual & Block-based & 86.0 & 30.0 & 161.2 & 57.6 & 164.1 \\
   & Dual & DCD & 83.1 & 27.3 & 155.8 & 55.9 & 165.3 \\
  \midrule

  \multirow{6}{*}{
  \begin{tabular}{c}
  Dream-v0- \\
  base-7B
  \end{tabular}}
  & None & Block-based & 129.7 & 120.2 & 169.9 & 64.9 & - \\
  & None & DCD & 113.2 & 123.5 & 162.0 & 51.3 & - \\
  & Prefix & Block-based & 132.8 & 130.1 & 160.8 & 65.8 & - \\
  & Prefix & DCD & 119.6 & 136.4 & 153.4 & 52.2 & - \\
  & Dual & Block-based & 132.5 & 138.3 & 163.2 & 68.1 & - \\
  & Dual & DCD & 121.2 & 141.0 & 149.6 & 53.6 & - \\
  \midrule

  \multirow{5}{*}{
  \begin{tabular}{c}
  Fast-dLLM \\
  -v2-7B
  \end{tabular}}
   & None & Sub-block-based & 157.0 & 130.1 & 216.4 & 108.3 & 253.4 \\
   & None & DCD & 143.9 & 124.1 & 196.7 & 92.5 & 243.9 \\
   & Dual & Sub-block-based & 157.9 & 136.9 & 224.5 & 112.7 & 258.2 \\
   & Dual & DCD & 143.2 & 121.8 & 198.0 & 94.5 & 246.4 \\
  \cmidrule(lr){2-8}
   & None & Block-based & 143.9 & 124.1 & 196.7 & 92.5 & 243.9 \\
   \midrule

 \multirow{1}{*}{
  \begin{tabular}{c}
  NBDiff
  \end{tabular}}
   & None & DCD & 4787.1 & 6185.1 & 7166.3 & 2573.3 & 3376.2 \\
 \bottomrule
  \end{tabular}
  }
\end{table*}

\begin{table*}[t]
  \centering
  \caption{Average forward length of each experiment.}
  \label{tab:fl}
  \resizebox{\textwidth}{!}{%
  \begin{tabular}{cccccccc}
  \toprule
  
  \textbf{Model} & \textbf{Cache} & \textbf{Decoding} & \textbf{Humaneval} & \textbf{MBPP} & \textbf{MATH500} & \textbf{GSM8K} & \textbf{IFEval} \\
  \midrule
  
  \multirow{6}{*}{%
  \begin{tabular}{c}
  LLaDA-8B \\
  -Instruct
  \end{tabular}}
   & None & Block-based & 672.3 & 1259.9 & 674.6 & 1540.3 & 571.1 \\
   & None & DCD & 672.5 & 1261.0 & 673.7 & 1540.9 & 571.1 \\
   & Prefix & Block-based & 317.4 & 416.5 & 347.0 & 504.3 & 354.9 \\
   & Prefix & DCD & 312.9 & 410.4 & 345.1 & 506.9 & 366.2 \\
   & Dual & Block-based & 86.1 & 134.7 & 83.0 & 172.2 & 56.9 \\
   & Dual & DCD & 88.8 & 139.0 & 81.5 & 180.8 & 72.8 \\
  \midrule
  
  \multirow{6}{*}{
  \begin{tabular}{c}
  Dream-v0- \\
  Instruct-7B
  \end{tabular}}
   & None & Block-based & 681.4 & 1234.7 & 667.9 & 1532.5 & 576.9 \\
   & None & DCD & 685.0 & 1241.2 & 668.1 & 1531.1 & 576.6 \\
   & Prefix & Block-based & 435.3 & 550.1 & 375.2 & 552.8 & 368.6 \\
   & Prefix & DCD & 436.4 & 572.6 & 373.3 & 560.2 & 383.3 \\
   & Dual & Block-based & 69.1 & 116.8 & 78.1 & 148.0 & 57.5 \\
   & Dual & DCD & 70.0 & 152.0 & 78.1 & 163.8 & 69.0 \\
  \midrule

  \multirow{6}{*}{
  \begin{tabular}{c}
  Dream-v0- \\
  base-7B
  \end{tabular}}
  & None & Block-based & 646.0 & 1203.3 & 646.9 & 1511.5 & - \\
  & None & DCD & 646.6 & 1203.1 & 647.7 & 1513.3 & - \\
  & Prefix & Block-based & 405.2 & 472.5 & 344.3 & 526.4 & - \\
  & Prefix & DCD & 415.4 & 470.3 & 344.2 & 558.9 & - \\
  & Dual & Block-based & 65.7 & 167.5 & 90.5 & 121.6 & - \\
  & Dual & DCD & 71.2 & 170.6 & 90.2 & 152.6 & - \\
  \midrule

  \multirow{5}{*}{
  \begin{tabular}{c}
  Fast-dLLM \\
  -v2-7B
  \end{tabular}}
   & None & Sub-block-based & 32.9 & 37.3 & 32.6 & 41.2 & 32.2 \\
   & None & DCD & 33.0 & 37.6 & 32.7 & 42.8 & 32.2 \\
   & Dual & Sub-block-based & 15.5 & 19.3 & 18.0 & 25.7 & 13.5 \\
   & Dual & DCD & 20.4 & 24.4 & 22.5 & 32.4 & 19.4 \\
  \cmidrule(lr){2-8}
   & None & Block-based & 33.0 & 37.6 & 32.7 & 42.8 & 32.2 \\
   \midrule
 \multirow{1}{*}{
  \begin{tabular}{c}
  NBDiff
  \end{tabular}}
   & None & DCD & 32.0 & 32.0 & 32.0 & 32.0 & 32.0 \\
 \bottomrule
  \end{tabular}
  }
\end{table*}

\end{document}